# Robust Multi-Image HDR Reconstruction for the Modulo Camera

Florian Lang    Tobias Plötz    Stefan Roth

Department of Computer Science, TU Darmstadt

**Abstract.** Photographing scenes with high dynamic range (HDR) poses great challenges to consumer cameras with their limited sensor bit depth. To address this, Zhao *et al.* recently proposed a novel sensor concept – the modulo camera – which captures the least significant bits of the recorded scene instead of going into saturation. Similar to conventional pipelines, HDR images can be reconstructed from multiple exposures, but significantly fewer images are needed than with a typical saturating sensor. While the concept is appealing, we show that the original reconstruction approach assumes noise-free measurements and quickly breaks down otherwise. To address this, we propose a novel reconstruction algorithm that is robust to image noise and produces significantly fewer artifacts. We theoretically analyze correctness as well as limitations, and show that our approach significantly outperforms the baseline on real data.

## 1 Introduction

Real world scenes often exhibit a significant dynamic range [17]. The intricate interplay between brightness and darkness, shadowy and sunny areas is often highly desirable from a photographer's standpoint. However, consumer cameras with image sensors that saturate when certain brightness levels are exceeded can only measure a significantly smaller dynamic range, *e.g.* 12 bit. When taking only a single image, the photographer faces the dilemma of losing detail either in the bright or in the dark parts of the scene; the whole scene cannot be captured in full detail. While various special sensors for HDR imaging have been developed, these are expensive [1], or sacrifice spatial [18] or intensity resolution [14].

Hence, various approaches aim to retain detail in the entire scene by reconstructing an HDR image from multiple captures, each with a different exposure time [4, 9, 10, 17]. As conventional image sensors saturate at some brightness level, the bit depth in bright parts of the scene is necessarily limited and the reconstruction may lead to artifacts. The *modulo camera* concept of Zhao *et al.* [22] aims to mitigate this using a novel, practical sensor that, instead of saturating, resets pixels to zero as soon as their maximal value is reached during the exposure. Hence, the least significant bits of the signal can be measured independently of its overall magnitude. This is in contrast to conventional cameras, which measure the signal correctly only up to the saturation level. To recover an HDR image from just a single modulo camera exposure, phase unwrapping techniques well-known in radar interferometry [8] or MRI [2] can be applied.

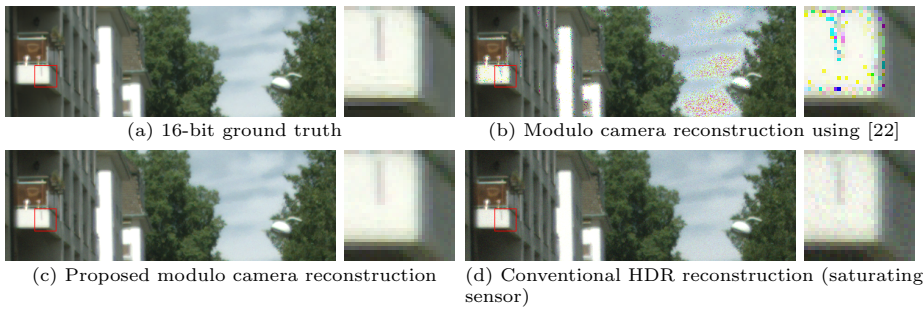

**Fig. 1.** Reconstruction of a 16-bit image from the Cityscapes dataset [3]. Multiple noisy 12-bit exposures of a modulo camera are simulated based on the ground truth in *(a)*. The approach of [22] produces visible outliers *(b)*. Our robust method *(c)* reconstructs the (noise-free) ground truth well. Reconstructing from the same number of exposures of a simulated conventional camera with saturation leads to much more noise *(d)*.

However, this requires the true image to be sufficiently smooth. For more complex and realistic scenes, [22] presents an approach for HDR reconstruction from multiple images. As in a conventional HDR pipeline, the exposure times of the captured images are chosen such that they measure different parts of the radiance range. The observed modulo values of each captured image are scaled by the exposure time and then iteratively combined into an estimate of the true HDR image. The benefit is that significantly fewer exposures are needed than with saturating sensors, making HDR imaging much more practical.

In this paper, we first consider the correctness of the reconstruction. As we will discuss, the original approach of Zhao *et al.* [22] provably works correctly, but only as long as there is no noise and exposure times can be set with perfect accuracy. Since the no-noise assumption is unrealistic, large areas of wrongly estimated pixels are produced when applied to images with typical noise levels. This is a significant impediment to the practical application of the modulo camera. To address this, we next introduce a novel, robust HDR reconstruction algorithm for modulo camera images. Analogous to the original approach, we calculate a simulated long exposure image and use its most significant bits. The crucial difference is that we additionally use the remainder of the simulated image and compare it to the actual values in the modulo capture. This way we can identify pixels for which the original algorithm would produce an incorrect reconstruction and correct these. Moreover, we make a number of theoretical contributions: First, we provide an analysis of correctness, showing that for a known noise distribution with certain properties, our algorithm reconstructs the true image even in the presence of noise. Moreover, we contribute explicit bounds on the optimal exposure times of the individual input images, which allow to assess the maximal measurable bit depth that can be recovered correctly with a chosen probability. Qualitative and quantitative experiments using realistic scenes show that our algorithm is robust against noise and consistently outperforms the original algorithm of [22]. Figure 1 shows a visual example.

## 2   Related Work

Capturing scenes with both bright and dark areas is challenging for consumer cameras, while the human visual system copes with such scenarios quite effortlessly. To address this gap, high dynamic range imaging aims to enable photographers to capture such scenes without loosing details. Significant research efforts have been dedicated to HDR imaging [17], resulting in techniques whose captured dynamic range even exceeds the capabilities of the human eye [12].

*Multiple exposures.* The perhaps most widely used family of HDR techniques is based on capturing several images with different exposure times (some under-, some overexposed) and combining these. Multishot methods can even be used with photographic film [4]. In general, they proceed by estimating the inverse of the camera response function and weighted averaging of the different images [19]. Since conventional cameras saturate at a certain level, long exposures do not add details to saturated regions. Consequently, many images with different exposure times are needed [9] to avoid quantization effects and artifacts [22]. Stumpfel *et al.* [20] report that seven conventional images are needed to capture the dynamic range of real life scenes, even with an elaborate selection of exposure times [10] and calibration. On the other hand, the modulo camera allows for an HDR reconstruction that exceeds the dynamic range of typical scenes from three modulo images alone [22] (assuming noise-free images). Other multishot HDR approaches avoid having to recover the response function [16]. All multishot methods are challenged by dynamic scenes as well as noise. Moving objects or camera ego-motion lead to ghosting artifacts in the reconstruction. Multiple approaches for ghost-free reconstruction have been proposed [7, 11, 13]. Here we focus on static scenes without camera movement, and leave the dynamic setting for future work.

*HDR sensors.* To increase the dynamic range without requiring multiple exposures, advanced sensors have been designed. They, for example, sacrifice spatial resolution for increased dynamic range using pixels of different size, effectively resulting in different exposures [18]. These have been realized in consumer cameras, *e.g.* the Fujifilm SuperCCD [6]. Using a high precision analog-digital converter can increase the dynamic range as well, while keeping the resolution. Another approach is to use thin-film on ASIC (TFA) sensors [1], which yield even more dynamic range. Adoption has been hindered by expensive production, and the sequential read-out for color images, requiring a three times longer exposure.

*Special sensors.* In contrast to conventional saturating sensors with increased bit depth, it is also possible to change the camera response function. Loose *et al.* [14] propose a sensor with a logarithmic response, which increases the dynamic range and allows for longer exposures without saturation. While the dynamic range increases, intensity resolution is reduced, especially for bright pixels. Another possibility is a sensor that measures gradients [21], which enables fine quantization and allows correcting saturated pixels. In contrast to these approaches, the modulo camera does not sacrifice spatial or intensity resolution. While the leading bits of the intensity values are lost, the details are preserved.

## 3   The Modulo Camera

Before introducing our approach, we formalize the properties of a conventional saturating sensor and review the modulo camera following [22]. For an ideal sensor with unbounded capacity, the observed image $I$ arises from the scene radiance $R \in \mathbb{R}_0^+$ and the exposure time $\tau > 0$ as

$$I(\tau R) = \lfloor \lambda(\tau R + \epsilon(\tau R)) \rfloor, \tag{1}$$

where $\lfloor \cdot \rfloor$ denotes the floor operation and $\lambda$ subsumes all multiplicative factors involved in the photon-to-digit conversion, *e.g.* the quantum efficiency, the analog amplification, and the analog-to-digital conversion factor. Inevitably, the observed image will be corrupted by noise $\epsilon$, which arises from the Poisson arrival process of photons and from the electronics involved in the imaging process. Following the literature [5], we model $\epsilon$ as intensity-dependent Gaussian noise

$$\epsilon(\tau R) \sim \mathcal{N}\big(0, \sigma^2(\tau R)\big) \qquad \text{with} \qquad \sigma^2(\tau R) = \beta_1 \tau R + \beta_2. \tag{2}$$

For simplicity, we assume *w. l. o. g.* that $0 \leq R < 2^K$ and $\lambda = 1$. Hence, an exposure time $\tau = 1$ will result in a digital image with bit depth $K$. In practice, pixel elements have limited capacity. Hence, a saturating camera $S(\cdot)$ with bit depth $L < K$ will clip the recorded signal at a maximal value of $2^L - 1$:

$$S(\tau R) = \min(I(\tau R), 2^L - 1). \tag{3}$$

Due to the clipping all image structure in high-intensity areas will be lost. In contrast, a modulo camera $M(\cdot)$ [22] with bit depth $L$ always retains the $L$ least significant bits, as the sensing element is reset once it hits the maximal value:

$$M(\tau R) = I(\tau R) \bmod 2^L = I(\tau R) - k \cdot 2^L. \tag{4}$$

The true intensities can be reconstructed from the modulo image by estimating the number of rollovers $k$, *i.e.* how many times each pixel has been reset.

*Multi-image reconstruction.* Given a series of exposure times $0 < \tau_1 < \ldots < \tau_n = 1$, we observe the corresponding modulo images $M_i = M(\tau_i R)$ with unknown rollover maps $k_i$. We iteratively reconstruct radiance maps $\tilde{R}_i, i = 1, \ldots, n$ approximating $R$, as well as rollover maps $\tilde{k}_i$ approximating $k_i$. We assume that $k_1 = 0$ for all pixels in $M_1$, *i.e.* the first image has no rollovers. This is equivalent to requiring that the first exposure time is sufficiently short with $\tau_1 \leq 2^{L-K}$. Hence, $\tilde{R}_1$ is obtained as the first modulo image $M_1$ divided by its exposure time

$$\tilde{R}_1 = \frac{M_1}{\tau_1}. \tag{5}$$

The original reconstruction algorithm of [22] now proceeds by recursively estimating the rollover map $\tilde{k}_i$ for the next exposure time and afterwards updating $\tilde{R}_i$. The rollover map for $M_i$ can be estimated from $\tilde{R}_{i-1}$ by scaling with the new exposure time $\tau_i$

$$\tilde{k}_i = \left\lfloor \frac{\tau_i \tilde{R}_{i-1}}{2^L} \right\rfloor. \tag{6}$$

Now we combine modulo image $M_i$ and rollover map $\tilde{k}_i$ into a refined estimate

$$\tilde{R}_i = \frac{\tilde{k}_i 2^L + M_i}{\tau_i}. \tag{7}$$

Recalling that $\tau_n = 1$, a $K$-bit high dynamic range image is finally obtained as

$$\tilde{I} = \lfloor \tau_n \tilde{R}_n \rfloor = \tilde{R}_n. \tag{8}$$

*Correctness.* We show in the supplemental material that this reconstruction is provably exact, *i.e.* $\tilde{I} = I(\tau_n R)$, if the fraction $2^L \frac{\tau_{i-1}}{\tau_i}$ is a positive integer *and* if the recorded images are not corrupted by noise, *i.e.* $I(\tau_i R) = \lfloor \tau_i R \rfloor$, for all $i = 1, \ldots, n$. However, even miniscule amounts of noise can already cause the estimation of the rollovers $\tilde{k}_i$ from Eq. (6) to be incorrect, leading to visible artifacts, see Fig. 1b. A simple numerical example showing the limitations of the original approach is $R = 256$, $L = 8$ and $\tau_1 = 0.4$, $\tau_2 = 1$, for which it is easy to see that $\tilde{I} = 0 \neq I(\tau_2 R)$ even without noise.

Moreover, we note that image noise cannot be removed beforehand, since the modulo operation does not commute with image filtering in general. This is true even for linear filters with rational weights $\omega_i \in \mathbb{Q}$, *i.e.* there exist $x_i \in \mathbb{Z}$ such that

$$\left(\sum_i \omega_i x_i\right) \bmod n \neq \left(\sum_i \omega_i (x_i \bmod n)\right) \bmod n. \tag{9}$$

## 4  Robust HDR Reconstruction

Estimating the rollover image with the original reconstruction algorithm in Eq. (6) is susceptible to noise as $\tilde{k}_i$ may not match the true rollover map $k_i$ that underlies the modulo observations $M_i$. We now present our algorithm that accounts for noise while reconstructing the scene radiance. In particular, we detect and correct for possible estimation errors $\Delta_i$ in $\tilde{k}_i$ such that

$$\hat{k}_i = \tilde{k}_i + \Delta_i, \qquad \Delta_i \in \mathbb{Z}. \tag{10}$$

Then we use $\hat{k}_i$ instead of $\tilde{k}_i$ for updating the radiance map as

$$\tilde{R}_i = \frac{\hat{k}_i 2^L + M_i}{\tau_i} = \frac{(\tilde{k}_i + \Delta_i) 2^L + M_i}{\tau_i}. \tag{11}$$

To find an optimal value for $\Delta_i$, we want the simulated image $\tilde{I}_i = \lfloor \tau_i \tilde{R}_i \rfloor$ given the current radiance to be close to the simulated image $\bar{I}_i = \lfloor \tau_i \tilde{R}_{i-1} \rfloor$ based on the previous radiance estimate. Since the noise governing the non-rounded image intensities is assumed Gaussian, we formulate the following least-squares problem:

$$\min_{\Delta_i} \mathcal{L}(\Delta_i) \doteq \min_{\Delta_i} \| \lfloor \tau_i \tilde{R}_i \rfloor - \lfloor \tau_i \tilde{R}_{i-1} \rfloor \|^2 \tag{12}$$

$$= \min_{\Delta_i} \| \tau_i \tilde{R}_i - \lfloor \tau_i \tilde{R}_{i-1} \rfloor \|^2 \tag{13}$$

$$= \min_{\Delta_i} \| (\tilde{k}_i + \Delta_i) 2^L + M_i - \lfloor \tau_i \tilde{R}_{i-1} \rfloor \|^2, \tag{14}$$

which holds as $\tau_i \tilde{R}_i \in \mathbb{Z}$ and we plugged in Eq. (11). Using Eq. (6), we have

$$\mathcal{L}(\Delta_i) = \left\| \Delta_i 2^L + M_i - \underbrace{\left( \lfloor \tau_i \tilde{R}_{i-1} \rfloor - \left\lfloor \frac{\tau_i \tilde{R}_{i-1}}{2^L} \right\rfloor 2^L \right)}_{\doteq D_i} \right\|^2. \tag{15}$$

Since $M_i$ and $D_i$ are modulo values, it holds that

$$-2^L < M_i - D_i < 2^L, \tag{16}$$

which implies that $\mathcal{L}(\Delta_i) > \mathcal{L}(0)$ for $|\Delta_i| \geq 2$. Thus, the cost $\mathcal{L}$ is minimized for $\Delta_i \in \{-1, 0, 1\}$. We can now read off the optimal value as

$$\Delta_i = \arg\min_{\Delta} \|\Delta 2^L + M_i - D_i\|^2 = \begin{cases} +1, & M_i - D_i < -2^{L-1} \\ -1, & M_i - D_i > +2^{L-1} \\ 0, & \text{else.} \end{cases} \tag{17}$$

This ensures that the reconstructed $\tilde{I}_i$ lies in the interval $[\bar{I}_i - 2^{L-1}, \bar{I}_i + 2^{L-1}]$.

*Optimality.* We now establish optimality conditions and show that our novel algorithm is indeed more robust than the original approach of [22]. To simplify notation, we write $I_i = I(\tau_i R)$ for the ideal image taken with an unbounded sensor. We then consider the compound noise between two images

$$e_i \doteq I_i - \frac{\tau_i}{\tau_{i-1}} I_{i-1} \tag{18}$$

$$= \epsilon(\tau_i R) - r(\tau_i R + \epsilon(\tau_i R)) - \frac{\tau_i}{\tau_{i-1}} \epsilon(\tau_{i-1} R) + \frac{\tau_i}{\tau_{i-1}} r(\tau_{i-1} R + \epsilon(\tau_{i-1} R)), \tag{19}$$

where $r(x) \doteq x - \lfloor x \rfloor$ is the rounding error with $0 \leq r(x) < 1$.

**Theorem 1.** *If $|e_i| \leq 2^{L-1} - 1$ for $i = 2, \ldots, n$, the robust algorithm reconstructs $I_i$ in every iteration correctly, i.e. $\tilde{I}_i \doteq \tau_i \tilde{R}_i = I_i$. Especially it holds that $\tilde{I}_n = \tilde{R}_n = I_n$ and hence the true radiance map is reconstructed as well as possible.*

*Proof sketch.* We proceed by induction. By assumption, the first modulo image has no rollovers, *i.e.* $I_1 = M_1$, and hence $I_1 = \tau_1 \tilde{R}_1$. The induction step proceeds by contradiction and relies on a case distinction on the possible values of $\Delta_i$ and whether the true $k_i$ is over- or underestimated. In particular, we show that

$$\tau_i \tilde{R}_i \neq I_i \implies |e_i| > 2^{L-1} - 1. \tag{20}$$

We refer to the supplemental material for details, but note that it is somewhat intuitive that Eq. (17) leads to a correct reconstruction of the rollover maps. Consequently, the robust algorithm leads to the radiance being reconstructed as well as possible even for substantial amounts of noise.

This is in contrast to the original reconstruction approach of [22], for which the robustness depends on the image values itself. For some image values, it tolerates $|e_i| = 2^L - 1$, while for others $|e_i| = 1$ already produces wrong results. As we have seen in the simple example above, even rounding errors can already cause the original approach to fail. Even if rounding errors could be eliminated by carefully choosing exposure times, a small amount of noise leads to similar failures.

## 5  Robust Capture Protocol

Theorem 1 shows that our reconstruction algorithm recovers the true intensities as long as the compound noise of imaging $\tau_i R$ and $\tau_{i-1} R$ is small enough. Since the noise and rounding error of $\tau_{i-1} R$ enter the definition of $e_i$ with a factor of $\tau_i/\tau_{i-1} > 1$ (Eq. 19), we can control the distribution of $e_i$ by choosing the ratio of exposure times $\tau_i/\tau_{i-1}$. For practical applications it is highly desirable to find an explicit exposure time schedule such that the number of exposures is minimized while guaranteeing correctness with high probability. Let us consider that in iteration $i$ we want to reconstruct a pixel correctly with some probability $p$. Then we aim to find the maximal $\tau_i$ given $\tau_{i-1}$ such that

$$\mathrm{P}\big[|e_i| \leq 2^{L-1} - 1\big] \geq p. \tag{21}$$

We can bound the probability on the left-hand side by the CDF of the absolute value of a Gaussian (for details see supplemental material)

$$\mathrm{P}\big[|e_i| \leq 2^{L-1} - 1\big] \geq \mathrm{P}\left[|\mathcal{N}(0, \sigma_i^2)| \leq 2^{L-1} - 1 - \frac{\tau_i}{\tau_{i-1}}\right] \tag{22}$$

with

$$\sigma_i^2 = \beta_1 \tau_i R \left(1 + \frac{\tau_i}{\tau_{i-1}}\right) + \beta_2 \left(1 + \frac{\tau_i^2}{\tau_{i-1}^2}\right). \tag{23}$$

This can be used to derive an upper bound on $\tau_i$ such that Eq. (21) is satisfied:

$$\tau_i \leq \tau_{i-1} \frac{-b + \sqrt{b^2 - 4ac}}{2a} \tag{24}$$

with

$$a = \beta_1 \tau_{i-1} R + \beta_2 - \big(\Phi^{-1}\big(\tfrac{1}{2} + \tfrac{1}{2}p\big)\big)^{-2} \tag{25}$$

$$b = \beta_1 \tau_{i-1} R + \big(\Phi^{-1}\big(\tfrac{1}{2} + \tfrac{1}{2}p\big)\big)^{-2}(2^L - 2) \tag{26}$$

$$c = \beta_2 - \big(\Phi^{-1}\big(\tfrac{1}{2} + \tfrac{1}{2}p\big)\big)^{-2}(2^{L-1} - 1)^2, \tag{27}$$

where $\Phi^{-1}$ is the inverse CDF of a standard normal distribution. We again refer to the supplemental material for details. The noise parameters $\beta_1, \beta_2$ can be estimated for a specific camera, *e.g.* with the technique of Foi *et al.* [5]. However, the scene radiance $R$ is needed in the calculation of $\tau_i$. We make a conservative guess and use the maximal value, *i.e.* $2^K$. That way every pixel is reconstructed correctly with probability at least $p$, while pixels with a lower intensity will be reconstructed with higher probability.

In general, if we always take the upper bound in Eq. (24) the resulting series will converge to a limiting exposure time $\tau^*$. We state an explicit formula for $\tau^*$ in the supplementary. Note, that the limiting exposure time can be greater than 1 and that it effectively allows us to characterize the *maximum bit depth* that can be resolved using a modulo camera with bit depth $L$ that, in contrast to [22], assumes realistic noise with parameters $\beta_1, \beta_2$. If $\beta_2$ is too large, $\tau^*$ will become complex-valued. In contrast, if $\beta_1$ is too large, we get $\tau^* < 1$, *i.e.* we cannot reconstruct $R$ with the required certainty.

## 6   Evaluation

We now analyze our robust reconstruction algorithm on real HDR data. We use the Cityscapes dataset [3], which provides 16-bit linear HDR images of road scenes. Since the images have already been debayered, we apply a color filter array to remove the interpolated pixels. Given the 16-bit images, we simulate the imaging process of an $L$-bit modulo camera as well as a conventional $L$-bit camera that goes into saturation. Note that we need to rely on simulation here, as modulo sensors have not been manufactured yet apart from early prototypes [22]. We choose the base parameters of the noise distribution as $10^{-8} \leq \beta_1 \leq 10^{-2}$ and $\beta_2 = 0.01\beta_1$, which are reasonable for consumer grade cameras [15]. In practice a certain camera has a fixed bit depth. In order to nevertheless compare the results across different bit depths for the same camera, we normalize the noise parameters such that an $L$-bit camera has $\beta_1^{(L)} = \alpha\beta_1$ and $\beta_2^{(L)} = \alpha^2\beta_2$, where $\alpha = 2^L - 1$. This implies that an $L$-bit camera collects $2^{K-L}$ times more photons to increment the intensity by one than the $K$-bit ground truth camera, thus improving the signal-to-noise ratio. This assumption is reasonable since cameras with a smaller bit depth can afford to integrate bigger capacitors instead.

*Maximal bit depth.* We first analyze the maximum bit depth that can be reconstructed and compare the proposed approach against the original reconstruction from [22]. We choose the exposure times according to Eq. (24) such that a pixel is reconstructed correctly with $p > 0.99$ in each iteration. While the effect of having a maximum of 1% of pixels potentially incorrect could be drastic when many iterations need to be made, this is a worst case estimate. In practice, many fewer errors occur, since the image is not bright everywhere. Moreover, we observe that choosing $p > 0.999$ or $0.9999$ did not yield significantly better results in practice, but required more images. Figure 2 shows how many bits can be reconstructed accurately from 2 and 5 exposures, respectively, as well as in the hypothetical case of infinitely many exposures, where the last exposure time is given by $\tau^*$. Depending on the bit depth of the camera and the noise characteristics, we achieve very high maximal exposure times and therefore a large dynamic range that can be recovered. For example, roughly 18 bits can be reconstructed with only 2 images, if $\beta_1 = 10^{-5}, \beta_2 = 10^{-7}$ and $L = 12$. Moreover, with just 5 exposures we already achieve a bit depth that is close to the theoretical maximum, meaning that the exposure time schedule given by Eq. (24) quickly converges to its limit. It is important to note that in the presence of realistic amounts of sensor noise, and even with our robust reconstruction algorithm, the modulo camera does not enable HDR images with an unbounded dynamic range, unlike what is claimed in [22]. Nevertheless, our derivation shows that even with realistic camera noise, significant gains in dynamic range can be achieved. Further improvements could potentially be made by considering the modulo images jointly, *c.f.* [8], instead of sequentially. We leave that for future work.

*HDR reconstruction.* We now compare the results from the original algorithm, our robust approach, and a simple HDR reconstruction method based on images from a saturating sensor, all with the same number of exposures. For the latter, we average over all images in which a pixel is neither under- nor oversaturated. We measure the quality of the reconstruction using the peak signal-to-noise ratio between the debayered and tone-mapped images: $\text{PSNR} \doteq 10 \log_{10}\left(\frac{\text{MAX}_I^2}{\text{MSE}}\right)$, where $\text{MAX}_I = 2^{16} - 1$ is the maximum intensity of the reconstructed image and MSE is the mean squared error. We determined the number of required exposures using the bound from Section 5. In the case of $\tau^* < 1$ we padded the first two exposure times to reach $\tau_n = 1$, for $\tau^* \ll 1$, i.e. extreme noise, we used 6 predefined exposures. Other schedules did not significantly affect the findings.

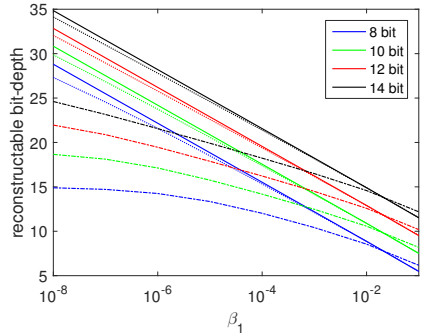

**Fig. 2.** Maximal bit depth that can be reconstructed from modulo images under realistic image noise using 2 (dot-dashed), 5 (dotted) and infinitely many (solid) images for varying bit depth and noise parameters. We set $\beta_2 = 0.01\beta_1$.

As we can see in Fig. 3 for low and moderate noise, the result of our robust algorithm is very close to the noisy ground truth. For strong noise the results become slightly suboptimal, since the calculated exposure times are too small, i.e. $\tau^* < 1$. The original modulo reconstruction approach of [22] and a simple HDR pipeline with a saturating sensor do much worse, since we are using as few exposures as possible. Especially for weak noise we often only need 2 images for our approach. Note, that the HDR pipeline with a saturating sensor combines a long-exposure image with a short-exposure image by appropriately upscaling the short-exposure image and subsequently averaging both images. In the upscaled image this effectively leads to an amplification of the noise parameter $\beta_2$ that is otherwise independent of the exposure time. This does not occur when reconstruction an HDR image using our approach as in every step only information from a single image is used to add detail to the reconstruction. Figure 4 shows an image from another dataset. We experience the same effects as in Fig. 1.

## 7 Conclusion

Modulo cameras have promised to be an interesting, practical alternative to conventional sensors that enable HDR reconstruction from a small number of exposures. However, as we have shown in this paper, the original reconstruction algorithm fails even in the noise-free case. Realistic image noise severely limits the attainable image quality. We proposed a novel, robust HDR reconstruction algorithm for images from a modulo camera and established clear criteria for its correctness. It can deal with significant amounts of noise and allows to assert stringent bounds on the probability of a successful reconstruction. We derived an optimal exposure time schedule for images with realistic noise and empirically showed that our robust algorithm performs very close to the ground truth, clearly outperforming the original modulo camera algorithm as well as standard HDR pipelines. Future work should consider extending the multi-image approach to dynamic scenes. Furthermore, an evaluation of a physical modulo camera in terms of the attainable noise level as well as production costs would be useful to determine the practicality of modulo sensors in consumer cameras.

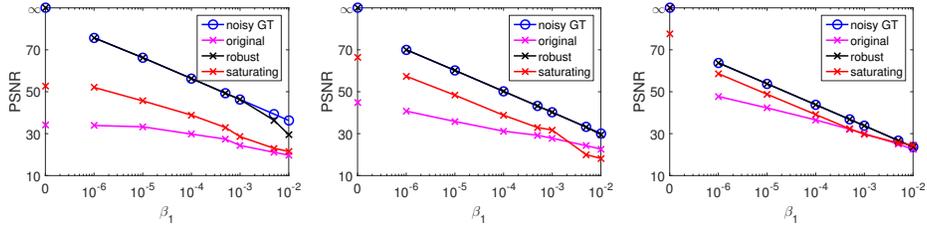

**Fig. 3.** PSNR of HDR reconstruction on the Cityscapes dataset with various methods for different noise levels ($\beta_2 = 0.01\beta_1$). *(from left to right)* 10, 12, 14 bit cameras. The overall noise increases with the bit depth since fewer photons are needed to increment the intensity value in a pixel site.

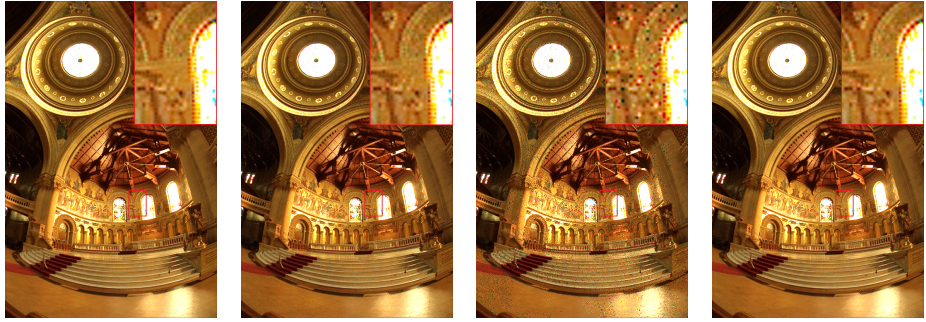

**Fig. 4.** HDR reconstruction of a scene in the Stanford Memorial Church. We use the HDR reconstruction of [4] from 18 exposures as ground truth for simulation. *(from left to right)* ground truth, conventional reconstruction using simulated saturating images, reconstruction from modulo images using the original algorithm, and using the proposed robust reconstruction approach. We simulated 12-bit cameras and added moderate amounts of noise ($\beta_1 = 10^{-3}, \beta_2 = 10^{-5}$).

*Acknowledgments.* The research leading to these results has received funding from the European Research Council under the European Union's Seventh Framework Programme (FP/2007–2013)/ERC Grant agreement No. 307942.

# Robust Multi-Image HDR Reconstruction for the Modulo Camera
## – Supplemental Material –

Florian Lang    Tobias Plötz    Stefan Roth

Department of Computer Science, TU Darmstadt

*Preface.* This supplementary material provides additional mathematical derivations. Specifically, we derive the explicit expression of the image differences $e_i$ (Eqs. 18 and 19), we prove Theorem 1 on the correctness of our robust algorithm, and prove that the original reconstruction algorithm from [22] is correct only for the case of no noise and carefully set exposure times. Moreover, we give a derivation of the exposure time schedule (Eq. 24) as well as the theoretical upper bound $\tau^*$ on the maximal achievable exposure time in the presence of image noise.

*Notation.* We quickly recap some of the notation used in the main paper. The image taken with a hypothetical, conventional $K$-bit camera and exposure time $\tau_i$ is defined as

$$I_i = \lfloor \tau_i R + \epsilon(\tau_i R) \rfloor = k_i 2^L + M_i. \tag{28}$$

The reconstructed image from our robust algorithm at time step $i$ is given by

$$\tilde{I}_i = \lfloor \tau_i \tilde{R}_i \rfloor = \tau_i \tilde{R}_i = (\tilde{k}_i + \Delta_i) 2^L + M_i. \tag{29}$$

Note that the lower $L$ bits of Eqs. (28) and (29) agree, because a $L$-bit modulo camera captures the $L$ least significant bits. For the following analysis it is convenient to define a simulated reconstruction $\bar{I}_i$ that uses the approximated radiance map $\tilde{R}_{i-1}$ from the previous time step instead of the updated radiance map $\tilde{R}_i$

$$\bar{I}_i = \lfloor \tau_i \tilde{R}_{i-1} \rfloor \doteq \tilde{k}_i 2^L + D_i. \tag{30}$$

Please note that $D_i$ is a proper modulo image, *i.e.* $0 \leq D_i < 2^L$. This follows from the definition of $\tilde{k}_i$ (Eq. 6):

$$\tilde{k}_i = \left\lfloor \frac{\tau_i \tilde{R}_{i-1}}{2^L} \right\rfloor. \tag{31}$$

## A  Derivation of $e_i$ (Eq. 19)

We now derive the explicit form of $e_i$ given by Eq. (19):

$$e_i = I_i - \frac{\tau_i}{\tau_{i-1}} I_{i-1} \tag{32}$$

$$= \lfloor \tau_i R + \epsilon(\tau_i R) \rfloor - \frac{\tau_i}{\tau_{i-1}} \lfloor \tau_{i-1} R + \epsilon(\tau_{i-1} R) \rfloor \tag{33}$$

$$= \tau_i R + \epsilon(\tau_i R) - r\bigl(\tau_i R + \epsilon(\tau_i R)\bigr) \tag{34}$$
$$\quad - \frac{\tau_i}{\tau_{i-1}} \Bigl( \tau_{i-1} R + \epsilon(\tau_{i-1} R) - r\bigl(\tau_{i-1} R + \epsilon(\tau_{i-1} R)\bigr) \Bigr)$$

$$= \epsilon(\tau_i R) - \frac{\tau_i}{\tau_{i-1}} \epsilon(\tau_{i-1} R) - \Bigl( \underbrace{r\bigl(\tau_i R + \epsilon(\tau_i R)\bigr)}_{\doteq r_i} - \frac{\tau_i}{\tau_{i-1}} \underbrace{r\bigl(\tau_{i-1} R + \epsilon(\tau_{i-1} R)\bigr)}_{\doteq r_{i-1}} \Bigr) \tag{35}$$

$$= \epsilon(\tau_i R) - \frac{\tau_i}{\tau_{i-1}} \epsilon(\tau_{i-1} R) - \Bigl( r_i - \frac{\tau_i}{\tau_{i-1}} r_{i-1} \Bigr). \tag{36}$$

Here, we model the effect of rounding as explicit rounding errors $r_i, r_{i-1}$ with $r(x) = x - \lfloor x \rfloor$ and $0 \leq r(x) < 1$. This allows us to work without floor functions in the following. As we can see, $e_i$ is given by the difference of both image noise terms as well as the difference of both rounding error terms.

## B  Proof of Theorem 1

We now prove the correctness of our robust reconstruction algorithm under the assumption that $|e_i| \leq 2^{L-1} - 1$. We use induction to show that $\tilde{I}_i = I_i$ for all $i = 1, \ldots, n$. The induction basis holds, since $\tilde{I}_1 = M_1 = I_1$ due to the initialization of our algorithm and the assumption that the exposure time $\tau_1$ is short enough such that the first image has no rollovers. For $i > 1$ we note that

$$\tilde{I}_i = I_i \tag{37}$$
$$\Leftrightarrow \quad (\tilde{k}_i + \Delta_i) 2^L + M_i = k_i 2^L + M_i \tag{38}$$
$$\Leftrightarrow \quad (\tilde{k}_i + \Delta_i) = k_i. \tag{39}$$

The least significant bits are equal by construction and it suffices to show that the leading bits, defined by the number of rollovers, are equal. To do that, we will proceed in two steps. First, we show that $e_i$ can be expressed in terms of $k_i$ and $\tilde{k}_i$ as

$$e_i = (k_i - \tilde{k}_i) 2^L + (M_i - D_i) - r', \tag{40}$$

where $r'$ is a rounding error term. Then, we show that assuming $(\tilde{k}_i + \Delta_i) \neq k_i$ implies that $|e_i| > 2^{L-1} - 1$, which contradicts our assumption that $|e_i| \leq 2^{L-1} - 1$. Hence, Eq. (39) must hold, which in turn implies Eq. (37).

**Step 1**

We derive $e_i$ in another way by looking at the difference between $\bar{I}_i$ and $I_i$, which we define as $\bar{e}_i$:

$$\bar{e}_i \doteq I_i - \bar{I}_i \tag{41}$$
$$= (k_i - \tilde{k}_i)2^L + (M_i - D_i) \tag{42}$$

where we used the definitions of $I$ (Eq. 28) and $\bar{I}$ (Eq. 30). We now show that $\bar{e}_i$ and $e_i$ are equal up to a rounding error, *i.e.* $\bar{e}_i = e_i + r'$, which concludes the first step (Eq. 40). By using the definitions of $I_i$ and $\bar{I}_i$ we have

$$\bar{e}_i = I_i - \bar{I}_i = \lfloor \tau_i R + \epsilon(\tau_i R) \rfloor - \lfloor \tau_i \tilde{R}_{i-1} \rfloor \tag{43}$$

$$= \tau_i R + \epsilon(\tau_i R) - r_i - \left\lfloor \frac{\tau_i}{\tau_{i-1}} \tau_{i-1} \tilde{R}_{i-1} \right\rfloor. \tag{44}$$

Now, we replace $\tilde{I}_{i-1} = \tau_{i-1} \tilde{R}_{i-1}$ with $I_{i-1} = \lfloor \tau_{i-1} R + \epsilon(\tau_{i-1} R) \rfloor$ by using the induction assumption

$$\bar{e}_i = \tau_i R + \epsilon(\tau_i R) - r_i - \left\lfloor \frac{\tau_i}{\tau_{i-1}} \lfloor \tau_{i-1} R + \epsilon(\tau_{i-1} R) \rfloor \right\rfloor \tag{45}$$

$$= \tau_i R + \epsilon(\tau_i R) - r_i - \left\lfloor \tau_i R + \frac{\tau_i}{\tau_{i-1}} \epsilon(\tau_{i-1} R) - \frac{\tau_i}{\tau_{i-1}} r_{i-1} \right\rfloor \tag{46}$$

$$= \tau_i R + \epsilon(\tau_i R) - r_i - \tau_i R - \frac{\tau_i}{\tau_{i-1}} \epsilon(\tau_{i-1} R) + \frac{\tau_i}{\tau_{i-1}} r_{i-1} \tag{47}$$

$$+ \underbrace{r(\tau_i R + \frac{\tau_i}{\tau_{i-1}} \epsilon(\tau_{i-1} R) - \frac{\tau_i}{\tau_{i-1}} r_{i-1})}_{=:r'} \tag{48}$$

$$= e_i + r'. \tag{49}$$

**Step 2**

We want to show the equality of rollovers (Eq. 39) by contradiction. Since $k_i, \tilde{k}_i, \Delta_i \in \mathbb{Z}$ it holds

$$(\tilde{k}_i + \Delta_i) \neq k_i \tag{50}$$

$$\Leftrightarrow \left( k_i - \tilde{k}_i \geq \Delta_i + 1 \quad \text{or} \quad k_i - \tilde{k}_i \leq \Delta_i - 1 \right). \tag{51}$$

We will show that

$$k_i - \tilde{k}_i \geq \Delta_i + 1 \implies e_i > 2^{L-1} - 1 \tag{52}$$

and

$$k_i - \tilde{k}_i \leq \Delta_i - 1 \implies e_i < -(2^{L-1} - 1) \tag{53}$$

leading to a contradiction since we assumed that $|e_i| \leq 2^{L-1} - 1$. To show Eqs. (52) and (53) we make a case distinction on $\Delta_i \in \{-1, 0, 1\}$.

**Case 1: $\Delta_i = 0$.** For $\Delta_i = 0$ we know from Eq. (17) that $-2^{L-1} \leq (M_i - D_i) \leq 2^{L-1}$. Now, for $k_i - \tilde{k}_i \geq \Delta_i + 1 = 1$ we have

$$e_i = (k_i - \tilde{k}_i)2^L + (M_i - D_i) - r' \tag{54}$$
$$> 2^L - 2^{L-1} - 1 \tag{55}$$
$$= 2^{L-1} - 1. \tag{56}$$

Analogously, for $k_i - \tilde{k}_i \leq \Delta_i - 1 = -1$ we have

$$e_i = (k_i - \tilde{k}_i)2^L + (M_i - D_i) - r' \tag{57}$$
$$\leq -2^L + 2^{L-1} - 0 \tag{58}$$
$$= -2^{L-1}. \tag{59}$$

**Case 2: $\Delta_i = 1$.** For $\Delta_i = 1$ we know that $-2^L + 1 \leq (M_i - D_i) \leq -2^{L-1} - 1$, with the left inequality following from $M_i$ and $D_i$ being modulo values and the right inequality following from Eq. (17). Now, for $k_i - \tilde{k}_i \geq \Delta_i + 1 = 2$ we have

$$e_i = (k_i - \tilde{k}_i)2^L + (M_i - D_i) - r' \tag{60}$$
$$> 2 \cdot 2^L - 2^L + 1 - 1 \tag{61}$$
$$= 2^L. \tag{62}$$

For the other case $k_i - \tilde{k}_i \leq \Delta_i - 1 = 0$ we have

$$e_i = (k_i - \tilde{k}_i)2^L + (M_i - D_i) - r' \tag{63}$$
$$\leq 0 - 2^{L-1} - 1 - 0 \tag{64}$$
$$= -2^{L-1} - 1. \tag{65}$$

**Case 3: $\Delta_i = -1$.** Similar to the last case, we can derive the two conditions

$$e_i > 2^{L-1} \quad \text{for} \quad k_i - \tilde{k}_i \geq \Delta_i + 1 = 0 \tag{66}$$
$$e_i \leq -2^L - 1 \quad \text{for} \quad k_i - \tilde{k}_i \leq \Delta_i - 1 = -2. \tag{67}$$

**Combination of cases.** Since we do not know a priori which of these three cases will occur, we need to pick the loosest condition. For $k_i - \tilde{k}_i \geq \Delta_i + 1$ we have

$$e_i > \min(2^{L-1} - 1, 2^L, 2^{L-1}) = 2^{L-1} - 1. \tag{68}$$

Similarly, for $k_i - \tilde{k}_i \leq \Delta_i - 1$ we have

$$e_i \leq \max(-2^{L-1}, -2^{L-1} - 1, -2^L - 1) = -2^{L-1}. \tag{69}$$

Hence, we see that in both cases even the loosest condition on $e_i$ violates our assumption that $|e_i| \leq 2^{L-1} - 1$. □

## C   Correctness of the Original Algorithm

Let us first present a simple yet not so obvious inequality concerning rounding errors.

**Lemma 1.** *For any $x \in \mathbb{R}$ and $c \in \mathbb{N}$ we have $r\left(\frac{x}{c}\right) \geq \frac{1}{c} r(x)$.*

*Proof.* We first apply simple transformations to the inequality:

$$r\left(\frac{x}{c}\right) \geq \frac{1}{c} r(x) \tag{70}$$

$$\Leftrightarrow \quad \frac{x}{c} - \left\lfloor \frac{x}{c} \right\rfloor \geq \frac{1}{c}(x - \lfloor x \rfloor) \tag{71}$$

$$\Leftrightarrow \quad \lfloor x \rfloor \geq c \left\lfloor \frac{x}{c} \right\rfloor. \tag{72}$$

The last inequality is true as for any $x, y \in \mathbb{R}$ it holds that $\lfloor x \rfloor + \lfloor y \rfloor \leq \lfloor x + y \rfloor$. Since $c \in \mathbb{N}$ we can apply this inequality $c$ times to obtain

$$c \left\lfloor \frac{x}{c} \right\rfloor \leq \left\lfloor c \frac{x}{c} \right\rfloor = \lfloor x \rfloor. \tag{73}$$

□

Now, we can prove the correctness of the original multishot reconstruction algorithm given noise-free observations and certain constraints on the exposure time schedule.

**Theorem 2.** *The reconstruction of the original algorithm of [22] is provably exact, i.e. $\tilde{I}_i = I_i$, if the fraction $2^L \frac{\tau_{i-1}}{\tau_i}$ is a positive integer, i.e. there exists $c \in \mathbb{N}$ with $c = 2^L \frac{\tau_{i-1}}{\tau_i}$, and if the recorded images are not corrupted by noise, i.e. $I_i = \lfloor \tau_i R \rfloor$, for all $i = 1, \ldots, n$.*

*Proof.* We proceed by induction. For $i = 1$ we have $\tilde{I}_1 = I_1$ by construction when assuming that the first image has no rollovers. For $i > 1$ we again show that $\tilde{k}_i = k_i$, which implies that $\tilde{I}_i = I_i$, c.f. Appendix B. We can now rewrite $\tilde{k}_i$ as

$$\tilde{k}_i = \left\lfloor \frac{\tau_i \tilde{R}_{i-1}}{2^L} \right\rfloor = \left\lfloor \frac{\tau_i}{\tau_{i-1}} \frac{\tau_{i-1} \tilde{R}_{i-1}}{2^L} \right\rfloor = \left\lfloor \frac{\tau_i}{2^L \tau_{i-1}} \lfloor \tau_{i-1} R \rfloor \right\rfloor, \tag{74}$$

where we used the induction hypothesis in the last step. Let us now further rewrite Eq. (74) by introducing explicit terms for the rounding error.

$$\tilde{k}_i = \left\lfloor \frac{\tau_i R}{2^L} - \frac{1}{c} r(\tau_{i-1} R) \right\rfloor = \left\lfloor \left\lfloor \frac{\tau_i R}{2^L} \right\rfloor + r\left(\frac{\tau_i R}{2^L}\right) - \frac{1}{c} r(\tau_{i-1} R) \right\rfloor \tag{75}$$

$$= \left\lfloor \left\lfloor \frac{\tau_i R}{2^L} \right\rfloor + r\left(\frac{\tau_{i-1} R}{c}\right) - \frac{1}{c} r(\tau_{i-1} R) \right\rfloor. \tag{76}$$

From the first equality it becomes clear that $\tilde{k}_i \leq k_i$, since

$$\tilde{k}_i = \left\lfloor \frac{\tau_i R}{2^L} - \frac{1}{c} r(\tau_{i-1} R) \right\rfloor \leq \left\lfloor \frac{\tau_i R}{2^L} \right\rfloor = k_i. \tag{77}$$

For the other direction, i.e. $\tilde{k}_i \geq k_i$, it suffices to show that

$$r\left(\frac{\tau_{i-1} R}{c}\right) \geq \frac{1}{c} r(\tau_{i-1} R). \tag{78}$$

This is true by invoking Lemma 1. Plugging the last inequality into Eq. (76) we see that

$$\tilde{k}_i = \left\lfloor \left\lfloor \frac{\tau_i R}{2^L} \right\rfloor + r\left(\frac{\tau_{i-1} R}{c}\right) - \frac{1}{c} r(\tau_{i-1} R) \right\rfloor \geq \left\lfloor \left\lfloor \frac{\tau_i R}{2^L} \right\rfloor \right\rfloor = k_i. \tag{79}$$

From Eqs. (77) and (79) we conclude that $\tilde{k}_i = k_i$. □

*Failure case.* We now give an example (assuming $L = 8$) to show that the original algorithm might fail even when the assumption on the $\tau_i$ is violated only slightly. Let $R = 256$, $\tau_1 = 0.4$ and $\tau_2 = 1$. Then, the true images are $I_1 = \lfloor 0.4R \rfloor = 102$ and $I_2 = \lfloor R \rfloor = 256$. The modulo images are $M_1 = 102$ and $M_2 = 0$. Executing the original algorithm yields

$$\tilde{R}_1 = \frac{M_1}{\tau_1} = 255 \tag{80}$$

$$\tilde{k}_2 = \left\lfloor \frac{255}{256} \right\rfloor = 0 \tag{81}$$

$$\tilde{R}_2 = \tilde{k}_2 2^8 + M_2 = 0 + 0 = 0. \tag{82}$$

Therefore, the final reconstruction $\tilde{I}_2 = 0 \neq 256 = I_2$ deviates from the true image by a wide margin.

## D  Derivation of Robust Exposure Time Schedule (Eq. 24)

We first simplify the distribution of image differences $e_i$. By inserting the intensity-dependent noise distributions Eq. (2) for $\epsilon$ in Eq. (36), we get

$$e_i \sim \mathcal{N}(0, \beta_1 \tau_i R + \beta_2) - \frac{\tau_i}{\tau_{i-1}} \mathcal{N}(0, \beta_1 \tau_{i-1} R + \beta_2) - r_i + \frac{\tau_i}{\tau_{i-1}} r_{i-1} \tag{83}$$

$$\sim \mathcal{N}\bigg(0, \underbrace{\beta_1 \tau_i R\big(1 + \tfrac{\tau_i}{\tau_{i-1}}\big) + \beta_2\big(1 + \tfrac{\tau_i^2}{\tau_{i-1}^2}\big)}_{=\sigma_i^2}\bigg) - r_i + \frac{\tau_i}{\tau_{i-1}} r_{i-1}. \tag{84}$$

According to Eq. (21) of the main paper, we want to find the maximal $\tau_i$ such that the reconstruction is correct with probability of at least $p$, i.e.

$$\mathrm{P}\big[|e_i| \leq 2^{L-1} - 1\big] \geq p. \tag{85}$$

We can now further bound the left-hand side of above equation by using the triangle inequality and exploiting that $0 \leq r_i < 1$ and $0 \leq r_{i-1} < 1$:

$$\mathrm{P}\left[|e_i| \leq 2^{L-1} - 1\right] = \mathrm{P}\left[\left|\mathcal{N}(0,\sigma_i^2) - r_i + \frac{\tau_i}{\tau_{i-1}}r_{i-1}\right| \leq 2^{L-1} - 1\right] \tag{86}$$

$$\geq \mathrm{P}\left[\left|\mathcal{N}(0,\sigma_i^2)\right| + \left|\frac{\tau_i}{\tau_{i-1}}r_{i-1} - r_i\right| \leq 2^{L-1} - 1\right] \tag{87}$$

$$\geq \mathrm{P}\left[\left|\mathcal{N}(0,\sigma_i^2)\right| \leq 2^{L-1} - 1 - \frac{\tau_i}{\tau_{i-1}}\right]. \tag{88}$$

In the last line we used the fact that $\tau_i/\tau_{i-1} > 1$ to bound $r_i$ and $r_{i-1}$ with 0 and 1, respectively. We now use the last expression to derive a slightly stricter criterion on the $\tau_i$ than Eq. (21):

$$\mathrm{P}\left[\left|\mathcal{N}(0,\sigma_i^2)\right| \leq 2^{L-1} - 1 - \frac{\tau_i}{\tau_{i-1}}\right] \geq p \tag{89}$$

$$\Leftrightarrow \mathrm{P}\left[\mathcal{N}(0,\sigma_i^2) \leq 2^{L-1} - 1 - \frac{\tau_i}{\tau_{i-1}}\right] \geq \tfrac{1}{2} + \tfrac{1}{2}p \tag{90}$$

$$\Leftrightarrow 2^{L-1} - 1 - \frac{\tau_i}{\tau_{i-1}} \geq \sigma_i \Phi^{-1}\left(\tfrac{1}{2} + \tfrac{1}{2}p\right), \tag{91}$$

where we used the symmetry of $\mathcal{N}(0,\sigma_i^2)$ and the fact that the inverse CDF of $\mathcal{N}(0,\sigma^2)$ is monotonically increasing and given by $\sigma\Phi^{-1}(p)$ with $\Phi^{-1}(p)$ being the inverse CDF of $\mathcal{N}(0,1)$. By denoting $l = 1/\Phi^{-1}\left(\tfrac{1}{2}+\tfrac{1}{2}p\right)$ we obtain the following condition on $\sigma_i^2$ such that the inequality (Eq. 91) is exact:

$$\sigma_i^2 = l^2(2^{L-1} - 1)^2 + l^2\left(\frac{\tau_i}{\tau_{i-1}}\right)^2 - 2l^2\frac{\tau_i}{\tau_{i-1}}(2^{L-1} - 1). \tag{92}$$

After plugging in the definition of $\sigma_i^2$ (Eq. 84) we can reorder terms in numerous simple steps that we omit for brevity. We finally arrive at a quadratic equation for the ratio of $\tau_i$ and $\tau_{i-1}$

$$a\left(\frac{\tau_i}{\tau_{i-1}}\right)^2 + b\frac{\tau_i}{\tau_{i-1}} + c = 0 \tag{93}$$

with

$$a = \beta_1 \tau_{i-1} R + \beta_2 - l^2 \tag{94}$$

$$b = \beta_1 \tau_{i-1} R + l^2(2^L - 2) \tag{95}$$

$$c = \beta_2 - l^2(2^{L-1} - 1)^2. \tag{96}$$

Solving this quadratic equation yields

$$\left(\frac{\tau_i}{\tau_{i-1}}\right)_{1,2} = \frac{-b \pm \sqrt{b^2 - 4ac}}{2a}. \tag{97}$$

Keeping only the positive exposure time yields the final recursion formula

$$\tau_i \leq \tau_{i-1} \frac{-b + \sqrt{b^2 - 4ac}}{2a}. \tag{98}$$

## E   Explicit Formula for $\tau^*$

To obtain a theoretical upper bound on the exposure time $\tau^*$ we set $\tau_i = \tau_{i-1} = \tau^*$, i.e. $\tau^*$ is the fixed point of the recursion formula (Eq. 98). This yields

$$\tau^* = \tau^* \frac{-b + \sqrt{b^2 - 4ac}}{2a} \tag{99}$$

$$\Leftrightarrow \quad 1 = \frac{-b + \sqrt{b^2 - 4ac}}{2a} \tag{100}$$

$$\Leftrightarrow \quad a(a + b + c) = 0 \tag{101}$$

$$\Leftrightarrow \quad \left((\beta_1 \tau^* R) + \beta_2 - l^2\right)\left[\left((\beta_1 \tau^* R) + \beta_2 - l^2\right)\right. \tag{102}$$

$$\left. + \left((\beta_1 \tau^* R) + l^2(2^L - 2)\right) + \left(\beta_2 - l^2(2^{L-1} - 1)^2\right)\right] = 0, \tag{103}$$

where we inserted Eqs. (94) to (96) at the last step. Again, a simple but tedious rearrangement of terms yields a quadratic equation in $\tau^*$:

$$(\tau^*)^2 A + \tau^* B + C = 0, \tag{104}$$

with

$$A = 2\beta_1^2 R^2 \tag{105}$$

$$B = 4\beta_1 \beta_2 R + (2^{L+1} - 6 - 2^{2L-2})\beta_1 R l^2 \tag{106}$$

$$C = 2\beta_2^2 + l^4(2^{2L-2} - 2^{L-1} + 4) + \beta_2 l^2(2^{L+1} - 2^{2L-2} - 2). \tag{107}$$

Now, we can solve Eq. (104) for $\tau^*$ yielding

$$\tau_{1,2}^* = \frac{-B \pm \sqrt{B^2 - 4AC}}{2A}. \tag{108}$$

Again we are only interested in a positive $\tau^*$ and therefore

$$\tau^* = \frac{-B + \sqrt{B^2 - 4AC}}{2A}. \tag{109}$$